%% file: StereoPIFU.tex
\documentclass[final]{cvpr}

\usepackage{times}
\usepackage{epsfig}
\usepackage{graphicx}
\usepackage{amsmath}
\usepackage{amssymb}
\usepackage{bm}

\usepackage{float}
\usepackage{cuted}
\usepackage{capt-of}
\usepackage{overpic}
\usepackage{multirow}
\usepackage{booktabs} 
\usepackage{lipsum}

\renewcommand{\thefootnote}{}

\usepackage[pagebackref=true,breaklinks=true,colorlinks,bookmarks=false]{hyperref}

\def\cvprPaperID{0707} 
\def\confYear{CVPR 2021}

\begin{document}


\title{StereoPIFu: Depth Aware Clothed Human Digitization via Stereo Vision} 
\author{\large Yang Hong\textsuperscript{1} \quad Juyong Zhang\textsuperscript{1}\thanks{Corresponding author} \quad Boyi Jiang\textsuperscript{1}  \quad Yudong Guo\textsuperscript{1} \quad Ligang Liu\textsuperscript{1} \quad Hujun Bao\textsuperscript{2} \vspace{1.5 mm}\\
{\normalsize \textsuperscript{1}University of Science and Technology of China \quad \textsuperscript{2}Zhejiang University}\\
{\tt\footnotesize \{hymath@mail., juyong@, jby1993@mail., gyd2011@mail., lgliu@\}ustc.edu.cn} \hspace{1 mm} {\tt\footnotesize bao@cad.zju.edu.cn} \\}

\twocolumn[{
\maketitle

\thispagestyle{empty}
\vspace*{-8mm}

\begin{center}
   \begin{overpic}
         [width=\linewidth]{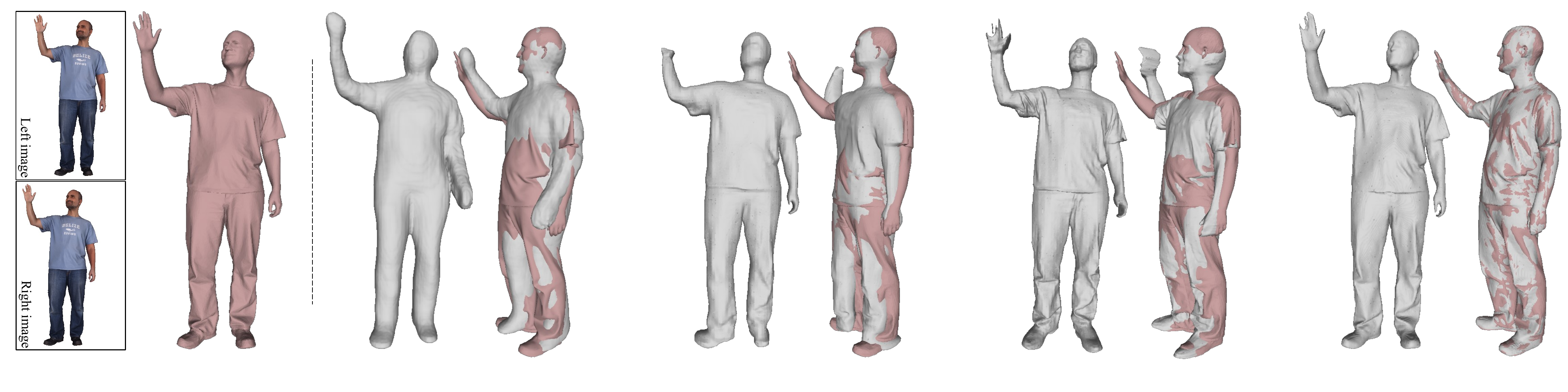}
         \put(3, 0.5){\footnotesize Input}
         \put(9.5, 0.5){\footnotesize Ground Truth}
         \put(25, 0.5){\footnotesize DeepHuman~\cite{zheng2019deephuman}}
         \put(42, 0.5){\footnotesize PIFu using 2-view images~\cite{saito2019pifu} }
         \put(67.5, 0.5){\footnotesize PIFuHD~\cite{saito2020pifuhd}}
         \put(91, 0.5){\footnotesize Ours}
   \end{overpic}
\end{center}
\vspace*{-2.5mm}
\captionof{figure}{Our proposed StereoPIFu could recover high-fidelity and depth-aware reconstruction of clothed human body. Compared with state-of-the-art methods, our reconstruction is more accurate and maintains correct relative positions of different parts of the human body.}
\label{fig:teaser}

\vspace*{5mm}

}]
{
  \renewcommand{\thefootnote}%
    {\fnsymbol{footnote}}
  \footnotetext[1]{Corresponding author}
}

\begin{abstract}
In this paper, we propose StereoPIFu, which integrates the geometric constraints of stereo vision with implicit function representation of PIFu, to recover the 3D shape of the clothed human from a pair of low-cost rectified images. First, we introduce the effective voxel-aligned features from a stereo vision-based network to enable depth-aware reconstruction. Moreover, the novel relative z-offset is employed to associate predicted high-fidelity human depth and occupancy inference, which helps restore fine-level surface details. Second, a network structure that fully utilizes the geometry information from the stereo images is designed to improve the human body reconstruction quality. Consequently, our StereoPIFu can naturally infer the human body's spatial location in camera space and maintain the correct relative position of different parts of the human body, which enables our method to capture human performance. Compared with previous works, our StereoPIFu significantly improves the robustness, completeness, and accuracy of the clothed human reconstruction, which is demonstrated by extensive experimental results.
\end{abstract}

\section{Introduction}
\input{Introduction.tex}

\section{Related Work}
\input{RelatedWork.tex}

\section{Method}
\input{Method.tex}

\section{Experiments}
\input{Experiments.tex}

\section{Discussion and Conclusion}
\input{Conclusion.tex}

{\small
\bibliographystyle{ieee_fullname}
\bibliography{egbib}
}

\end{document}

%% file: Introduction.tex
Human digitization is the key to many applications like AR/VR, virtual try-on, holographic communication, film/game production~\etc. While high-fidelity and geometric detail preserved 3D human digitalization can be achieved with high-end acquisition equipment and well-designed capture environment~\cite{collet2015high,guo2019relightables}, it is not suitable for general consumers. Recently, with the popularization of consumer-level acquisition devices, human digitization with simple inputs becomes a hot research topic in the field of 3D computer vision and computer graphics.

Many methods have been proposed for clothed human reconstruction from simple inputs (e.g., single image). Among them, some methods reconstruct the 3D human body with the help of parametric models~\cite{bogo2016keep,kanazawa2018end,kocabas2020vibe,kolotouros2019learning,SMPL:2015, pavlakos2019expressive,jiang2020disentangled}. However, parametric models are mainly used for reconstructing naked body and can not deal with topology changes. To solve these problems, implicit function based representations have been recently introduced~\cite{huang2018deep, park2019deepsdf,mescheder2019occupancy,saito2019pifu,saito2020pifuhd,chibane2020implicit,bhatnagar2020combining,gropp2020implicit,he2020geo}.  Representatively, Saito~\etal~\cite{saito2019pifu} proposed Pixel-Aligned Implicit Function (PIFu), which performs implicit function prediction based on the $z$-value of a 3D query point and its projected 2D image feature. PIFu is memory-efficient and can generate a plausible surface with a single image. Later, PIFuHD~\cite{saito2020pifuhd} further improves the results of PIFu in the aspect of fine-level geometric details recovery with the aid of predicted normal maps and higher resolution. However, like other single image-based approaches, PIFu can not predict the precise spatial location and suffers from depth ambiguity, which results in inconsistent results noticeable from different perspectives. Besides, PIFu related methods are prone to generate wrong structures like broken limbs.

A natural way to resolve the depth ambiguity is to take multi-view images as input. However, existing approaches for 3D human digitization rarely fully utilize the rich geometric relationships endowed by stereo vision. For example, Gilbert~\etal~\cite{gilbert2018volumetric} only exploits visual hull extracted from multi-view silhouettes, and PIFu-like methods~\cite{huang2018deep,saito2019pifu} aggregate the projected pixel features from multi-view images with a pooling layer for each query point. As shown in Fig.~\ref{fig:teaser}, this simple operation fails to perceive the depth information. On the other hand, some deep learning-based depth estimation methods explore geometric correlations between multi-view images and obtain satisfying results~\cite{kendall2017end,chang2018pyramid,khamis2018stereonet,chabra2019stereodrnet,zhang2019ga,xu2020aanet}. The key to superior performance is the constructed cost volume feature, which encodes the correlation between a pixel and its candidate pixels in another view. Generally, these methods predict disparity, namely the horizontal displacement between a pair of corresponding pixels from rectified stereo pair images. 
The pixel's predicted disparity can be obtained by calculating the weighted average of these candidate disparities, and the predicted depth can be recovered via triangulation~\cite{khamis2018stereonet, chang2018pyramid, xu2020aanet}.

To overcome the limitations of classic PIFu-like methods mentioned in the above, we propose StereoPIFu (Stereo Vision-Based Pixel-Aligned Implicit Function) for depth-aware clothed humans digitization. Compared with PIFu, StereoPIFu has two novel designs to improve its representation ability. First, we introduce additional novel voxel-aligned features as input to the implicit function. Different from the voxel-aligned features of previous methods~\cite{he2020geo, chibane2020implicit}, our voxel-aligned features are extracted from volume data widely used in stereo vision-based depth estimation networks. Specifically, we construct two volume data from the feature maps of the image pair. The volumes define a spatial grid and for a 3D query point, we can do trilinear interpolation on its neighboring voxels in the volumes to generate voxel-aligned features. The features contain rich geometric information, indicating the correlation between the query point and the underlying surface. 
With voxel-aligned features as input, StereoPIFu can predict humans' spatial location in camera space without the need to normalize human shape into a canonical space like PIFu. 
Second, we introduce human shape priors to guide the representation. Specifically, a high-quality human depth map is obtained based on the above-mentioned volume data. Moreover, the relative $z$-offset between the query point and its projected pixel's predicted depth is added as another input of the implicit function. Compared with using the absolute $z$-value, relative $z$-offset can generate more realistic details. More importantly, the predicted depth map provides complete surface constraints for the human body and effectively eliminates broken limbs in reconstructed results. With the above novel designs, StereoPIFu can recover high-fidelity geometric details and accurate geometric shape of human body. In summary, the paper includes the following contributions:

\begin{itemize}
    \item We propose StereoPIFu, a novel implicit representation integrating stereo vision to PIFu representation, which makes full use of binocular images and enables high-quality depth-aware reconstruction of the clothed human body.
    \item We utilize the novel well-designed voxel-aligned features and predicted depth map to help occupancy inference. The geometric correlation contained in the voxel-aligned features and depth priors significantly improve the robustness, completeness, and accuracy of clothed human reconstruction.
\end{itemize}

%% file: RelatedWork.tex
\noindent {\bf{Reconstruction from Single Image.}} Different strategies have been proposed to recover 3D human body shape from a single image. A common way is to fit the input image by a parametric model (e.g., SMPL~\cite{SMPL:2015}) via 2D landmarks and silhouette~\cite{bogo2016keep, kanazawa2018end,pavlakos2019expressive,kocabas2020vibe, jiang2020disentangled}. However, these methods can only recover the naked body's shape due to the limited representation ability of the parametric model. Several methods~\cite{alldieck2018video, alldieck2018detailed, alldieck19cvpr, alldieck2019tex2shape} add a displacement on each vertex to represent fine-scale details. Bhatnagar~\etal~\cite{bhatnagar2019multi} and Jiang~\etal~\cite{jiang2020bcnet} additionally regress the clothes via parametric models. These methods have improved the quality but are still unable to recover high-fidelity geometry shape.

Several methods~\cite{varol2018bodynet, zheng2019deephuman} take 3D CNN to regress volumetric representation of human bodies. However, they can not recover geometric details due to the large memory consumption of 3D CNN. Instead of regressing a volume with a fixed resolution, Saito~\etal~\cite{saito2019pifu} propose Pixel-Aligned Implicit Function~(PIFu) representation, which is memory-efficient and can predict the occupancy of any 3D point. However, it may generate incorrect body structures and struggle to recover fine-scale geometric details. To this end, PIFuHD~\cite{saito2020pifuhd} improves PIFu with additionally extracted high-resolution features and predicted normal maps. Huang~\etal~\cite{huang2020arch} and Zheng~\etal~\cite{zheng2020pamir} take a feature related to the parametric human model as the input of the implicit function to improve the stability of prediction. He~\etal~\cite{he2020geo} expands the pixel-aligned feature with geometry-aligned shape features, which serves as a shape prior for the reconstruction. Gabeur~\etal~\cite{gabeur2019moulding} convert this problem as depth prediction of the front- and back-side, and the complete shape is obtained by merging the recovered point clouds together. However, these methods suffer from depth ambiguity due to the inherent nature of single image input.

\noindent {\bf{Reconstruction from Multi-view Images.}} Previous studies extract shape cues from the silhouette, stereo, and shading, to recover geometric shape from multi-view inputs~\cite{song2010volumetric, collet2015high, xu2017shading, alldieck2018video, guo2019relightables}. The high-quality 3D shape can be reconstructed with hundreds of cameras~\cite{collet2015high, guo2019relightables,furukawa2009accurate, vzbontar2016stereo}, but this hardware configuration is inaccessible to general consumers due to its special equipment and complexity. Gilbert~\etal~\cite{gilbert2018volumetric} predicts the high-fidelity volume of the human shape utilizing a coarse visual hull generated by sparse view silhouettes with a 3D convolutional autoencoder network. Alldieck~\etal~\cite{alldieck2018video} take a video of humans slowly rotating as input and exploit the silhouette of frames to optimize the SMPL+D representation. Alldieck~\etal~\cite{alldieck19cvpr} uses a network to substitute the optimization, which leads to faster inference and simpler input (1-8 frames). However, these methods are hard to recover fine-level surface details due to their limited geometric representation abilities. Several works~\cite{huang2018deep, saito2019pifu} aggregate multi-view pixel-aligned features with a pooling layer to help reconstruction. Although improved results can be obtained, the simple aggregation with the pooling strategy does not make full use of the information endowed by multi-view images.

\noindent {\bf{Depth Estimation from Stereo Images.}} Depth estimation from stereo images has been studied extensively for decades. 
Although traditional methods have made significant progress~\cite{bleyer2011patchmatch, hosni2012fast}, they still suffer from the edge-fattening~\cite{scharstein2002taxonomy, min2011revisit} and poor performance in challenging situations like textureless regions. Recently, deep learning-based methods have been proposed to alleviate these problems~\cite{mayer2016large, kendall2017end,chang2018pyramid,khamis2018stereonet,chabra2019stereodrnet,zhang2019ga,xu2020aanet}. DispNet~\cite{mayer2016large} constructs a large-scale synthetic dataset, Scene Flow, and builds the first end-to-end trainable framework for disparity prediction. GC-Net~\cite{kendall2017end} concatenates the left and right images' feature and applies 3D convolutions to aggregate the resulting 4D feature cost volume. PSMNet~\cite{chang2018pyramid} further improves the accuracy by using a stacked hourglass block~\cite{newell2016stacked} with more 3D convolutional layers. Several methods~\cite{zhang2019ga, xu2020aanet} design novel modules to reduce the memory consumption and can obtain competitive performance with other state-of-the-art methods. However, these methods can only recover the shape of the human body's visible part as they do not utilize any human body shape prior.

%% file: Method.tex
\begin{figure*}
    \centering
    \includegraphics[width=\textwidth]{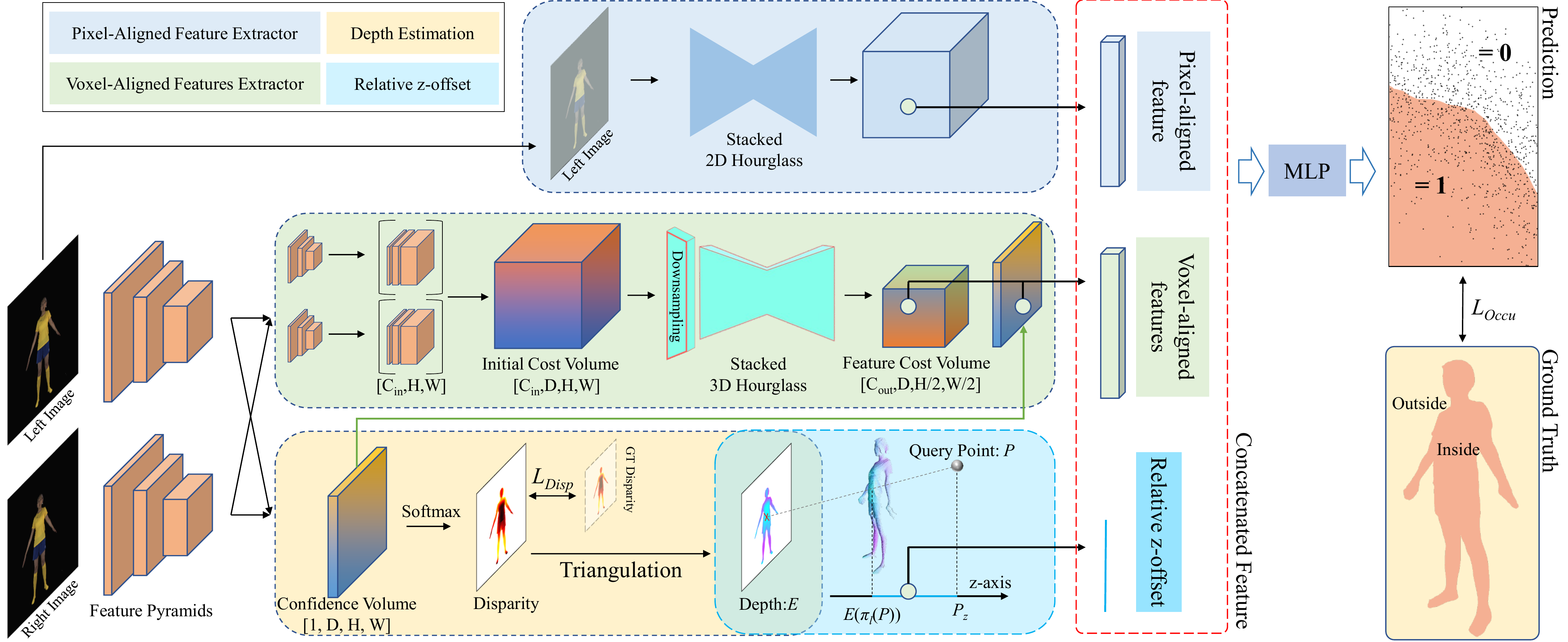}
    \caption{Overview of our StereoPIFu pipeline. Given a stereo pair, for a query point $P$, its pixel-aligned feature, voxel-aligned features, and relative z-offset are constructed. These features encode the information about whether $P$ is inside the underlying surface or not and are used for inferring the occupancy of $P$ by the MLP.}
    \label{fig:pipeline}
\end{figure*}

In this paper, we aim to perform depth-aware reconstruction for the clothed human. To this end, we propose StereoPIFu, which takes a pair of stereo images as input. Similar to PIFu~\cite{saito2019pifu}, we learn an implicit function that indicates whether a 3D point is inside the human body or not. Compared with PIFu, we additionally add the self-designed voxel-aligned features as input, which effectively improves our method's expression ability and enables depth-aware reconstruction of the clothed human body. Moreover, the predicted high-fidelity depth map further helps recover the fine-level geometric details of the visible part. An overview of our method is shown in Fig.~\ref{fig:pipeline}, and we will give the algorithm details for each part in the following.

\subsection{Stereo Vision-based PIFu}
StereoPIFu takes a pair of rectified color images, $\bm{I}_l$ and $\bm{I}_r$ as input, and can be formulated as:
\begin{equation}
    \begin{aligned}
        f(F_l(\pi_l(P)), \Phi(P), \Psi(P), \psi_t(Z_{E}(P)))  \mapsto s \in [0, 1],
    \end{aligned}
    \label{eq:functioninputs}
\end{equation}
where $f$ is the implicit function represented by a multi-layer perceptron (MLP). It infers a continuous scalar $s\in[0,1]$ to indicate the occupancy of any 3D query point $P$ in the left camera space. $\pi$ represents the camera projection. Its subscripts indicate its left and right image counterparts. $F_l$ is the feature map extracted from the left image and $F_l(\pi_l(P))$ denotes pixel-aligned feature in PIFu~\cite{saito2019pifu}. $\Phi(P)$ and $\Psi(P)$ are voxel-aligned features of $P$, and they will be described in detail later. We denote $E$ as the predicted depth map corresponding to $\bm{I}_l$, and the relative $z$-offset $Z_{E}(P)$ of $P$ relative to $E$ is defined as:
\begin{equation}
    \begin{aligned}
        Z_{E}(P) = P_z - E(\pi_l(P)),
    \end{aligned}
\end{equation}
where $P_z$ is the $z$-coordinate of $P$, and $E(\pi_l(P)$ is the depth value of $\pi_l(P)$ in $E$. Compared with PIFu~\cite{saito2019pifu}, we additionally input more variables to predict the occupancy value $s$ of $P$. We will introduce them one by one and explain the reasons for our design. 

\noindent{\bf{Pixel-Aligned Feature $\bm{F_l(\pi_l(P))}$.}} The pixel-aligned feature was first introduced in PIFu~\cite{saito2019pifu} and widely applied in the later PIFu-like methods \cite{saito2020pifuhd,huang2020arch,he2020geo,li2020robust,zheng2020pamir}. First, an image encoder (\eg, hourglass in \cite{saito2019pifu}) is used to extract feature map $F_l$ from $\bm{I}_l$. Generally, the image encoder is specially designed to have a large receptive field to support overall perception and consistent depth inference. For a given 3D query point $P$, we compute its pixel-aligned feature $F_l(\pi_l(P))$ by bilinearly interpolating $F_l$ at $\pi_l(P)$. Instead of using a global feature~\cite{park2019deepsdf}, the pixel-aligned feature encodes the local detail information contained in the image and results in a more high-fidelity reconstruction.

\noindent{\bf{Voxel-Aligned Features $\bm{\Phi(P)}$, $\bm{\Psi(P)}$.}} 
Although the pixel-aligned feature leads to more surface details, it cannot perceive the absolute depth information and struggles to guarantee correct relative positions of various body parts. In addition to the pixel-aligned feature, PIFu~\cite{saito2019pifu} further uses the query points' $z$-coordinates to distinguish their occupancy values along the ray. Therefore, it needs to normalize all training data to a fixed bounding box, making PIFu unable to restore the human body's spatial location. To this end, we specially design the novel voxel-aligned features, which contain rich spatial and geometric information. Based on these features, both the spatial location and the relative position of various body parts can be recovered.

Our observation is that the volume data, widely used in stereo vision-based depth estimation networks, naturally contains spatial information. For a given 3D query point $P$, we define its voxel-aligned features as the trilinear interpolation of these volume data depending on its spatial position. Specifically, we use AANet+, recently introduced in \cite{xu2020aanet}, to predict the depth map $E$ corresponding to $\bm{I}_l$. 
Its structure is shown in the lower-left part of Fig.~\ref{fig:pipeline}. First, a shared feature extractor is used to extract the downsampled feature pyramids.
Then, given the maximum disparity range $D$, an intermediate variable, i.e., the confidence volume $\Psi$, is constructed. It is a 4D tensor with size $[1, D, H, W]$ and defines a scalar field in 3D space~(please refer to~\cite{xu2020aanet} for details). For any element of $\Psi$, its index coordinate~$(d, i, j)$ corresponds to a 3D point in the space, and its value describes the probability of whether the point lies on the visible part of the underlying surface. Finally, the predicted disparity ${d}^{Pred}(i,j)$ for a pixel $(i,j)$ in the left image $\bm{I}_l$ is a combination of all the disparity values weighted by their corresponding confidence value in $\Psi$, and it is expressed as:
\begin{equation}
    {d}^{Pred}(i,j) = \sum_{d=0}^{D-1} d \times \Psi(d, i, j).
\end{equation}
The depth value can be computed by $\frac {bk}{{d}^{Pred}(i, j)}~$\cite{szeliski2010computer}, where $b$ is the baseline of stereo images and $k$ is the focal length.

As the confidence volume mainly encodes the visible part of the underlying surfaces, we introduce another feature cost volume to further encode the 3D space containing the entire underlying surface. As shown in the voxel-aligned features extractor of Fig.~\ref{fig:pipeline}, we first upsample the multi-scale features of two feature pyramids to the same size. Then, we concatenate them together to form two 3D tensors with size $[C_{in}, H, W]$, denoted as $\mathcal{F}_l$ and $\mathcal{F}_r$, respectively. Next, we follow Khamis~\etal~\cite{khamis2018stereonet} to construct the initial feature cost volume $\Phi'$ with size $[C_{in}, D, H, W]$ from $\mathcal{F}_l$ and $\mathcal{F}_r$, by computing the differences between a feature in $\mathcal{F}_l$ and its corresponding features with some disparities in $\mathcal{F}_r$, i.e.:
\begin{equation}
    \Phi'(:, d, i, j) = \mathcal{F}_l(:, i, j) - \mathcal{F}_r(:, i, j - d).
\end{equation}
Based on $\Phi'$, we use a variation of the stacked 3D hourglass network in~\cite{chang2018pyramid} to aggregate context across the spatial and disparity domain to generate the feature cost volume $\Phi$. 
Specifically, we add a downsampling layer at the beginning of the network for reducing memory consumption. All 3D deconvolutions are replaced by the combination of upsampling and convolution for eliminating the checkerboard artifacts~\cite{odena2016deconvolution}. 
The repeated bottom-up and top-down architecture of the hourglass network extends the 3D receptive field, making features in $\Phi$ more consistent and robust. 
Like the above-mentioned confidence volume $\Psi$, $\Phi$ naturally defines a vector field in 3D space, the index coordinates of each element of $\Phi$ corresponds to a 3D point, and its feature vector encodes the information whether the 3D point lies on the underlying surface, including the occluded parts.

For a given query point $P$, its voxel-aligned features consist of cost feature $\Phi(P)$ and confidence value $\Psi(P)$. They are obtained by trilinearly interpolating $\Phi$ and $\Psi$ at the coordinate $\big( \pi_{l}(P)_x - \pi_{r}(P)_x, \pi_{l}(P)_y, \pi_{l}(P)_x \big)$, respectively.

\noindent{\bf{Relative z-offset $\bm{Z_{E}(P)}$.}} 
As stated in PIFuHD~\cite{saito2020pifuhd}, the pixel-aligned feature with low resolution is hard to encode fine-level surface details. The above-mentioned voxel-aligned features also suffer a similar problem.
Fortunately, AANet+ can provide a high-resolution depth map. Thus the relative z-offset naturally encodes the fine-level geometric information. Moreover, thanks to the human shape prior from our constructed large-scale human body dataset~(see Sec.~\ref{sec:datasets}), the AANet+ retrained by the dataset can generate a high-quality, complete, and consistent human depth map. Furthermore, the relative z-offset can exploit the human priors from predicted depth to guide occupancy inference. Specifically, the relative z-offset endows the query points close to the underlying surface with a similar relative z-offset value, making the network easier to train and enabling our StereoPIFu to generate fine-level surface details and avoids broken limbs. Therefore, the relative z-offset actually acts as a bridge between the predicted depth and occupancy prediction. 

For some cases such as a hand in front of the torso, there will be some discontinuous regions in the predicted depth map $E$. In these cases, $Z_{E}(P)$ of the occluded query points in the back will change discontinuously, resulting in unnatural geometric copy in the occluded region, as shown in Fig.~\ref{fig:ablationstudy_psi}. To tackle this issue, we design a transformation function $\psi_t(x)$ to normalize $Z_{E}(P)$ to the interval $(-1.0, 1.0)$:

\begin{equation}
    \psi_t(x) = \frac{2.0}{1.0 + \exp(-t\cdot x)} - 1.0 ,
\end{equation}
where $t$ is a hyperparameter. For query points far away from the predicted depth, $\psi_t(Z_{E}(P))$ will be close to $1.0$ or $-1.0$, which effectively eliminates the geometric copy of the occluded part of human body and helps to produce more natural reconstruction results, see~Sec.~\ref{sec:evaluations}.

\subsection{Loss Function}
With our constructed dataset (see Sec.~\ref{sec:datasets}), our StereoPIFu is trained with ground truth as supervision. In the following, we give details of the loss terms.

\noindent {\bf{Loss on Depth Estimation.}} We adopt the same multi-scale loss function as \cite{xu2020aanet} to retrain AANet+, i.e., depth estimation module of our StereoPIFu, especially for human body type images, 
\begin{equation}
    L_{Disp} = \sum_{\omega}\lambda_{\omega} \cdot \mathcal{L}(d_{\omega}^{GT}(p), {d}_{\omega}^{Pred}(p))
    \label{equ:stereoloss}
\end{equation}
where ${d}_{\omega}^{Pred}(p)$ is $\omega$-scale predicted disparity in pixel $p$, and $d_{\omega}^{GT}(p)$ is the corresponding ground truth value. $\mathcal{L}$ is the smooth L1 loss and $\lambda_{\omega}$ is the weight to balance different terms. 

\noindent {\bf{Loss on Implicit Function.}} We adopt the same sampling strategy with \cite{saito2019pifu,saito2020pifuhd} to generate sampling points for training our implicit function. Differently, we do not normalize our training data to a canonical space and they are accompanied by random spatial locations (see Sec.~\ref{sec:datasets}). The Binary Cross Entropy (BCE) loss is used for occupancy prediction.
\begin{equation}
    \begin{aligned}
        L_{Occu} = \sum_{P\in \mathcal{S}} &f^{GT}(P)\cdot \log({f}^{Pred}(P)) \\
                                            &+ (1 - f^{GT}(P))\cdot \log(1 - {f}^{Pred}(P))
        \label{equ:occupancyloss}
    \end{aligned}
\end{equation}
where $\mathcal{S}$ denotes the set of sampling points, ${f}^{Pred}(P)$ is the predicted occupancy value for sampling point $P$ and $f^{GT}(P)$ is the corresponding ground truth occupancy value. The pixel-aligned feature extractor, the voxel-aligned features extractor, and the MLP of Fig.~\ref{fig:pipeline} are trained using $L_{Occu}$.

%% file: Experiments.tex
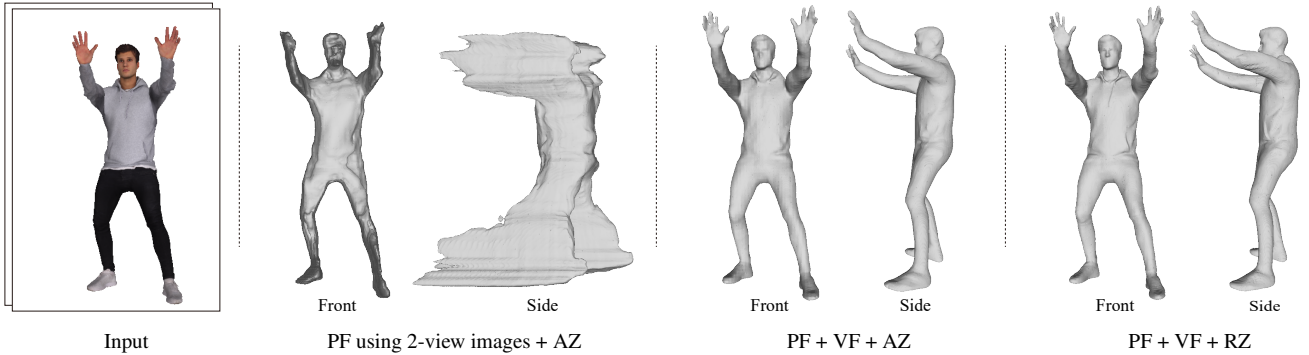
\begin{figure*}[thp]
    \centering
    \begin{overpic}
        [width=\linewidth]{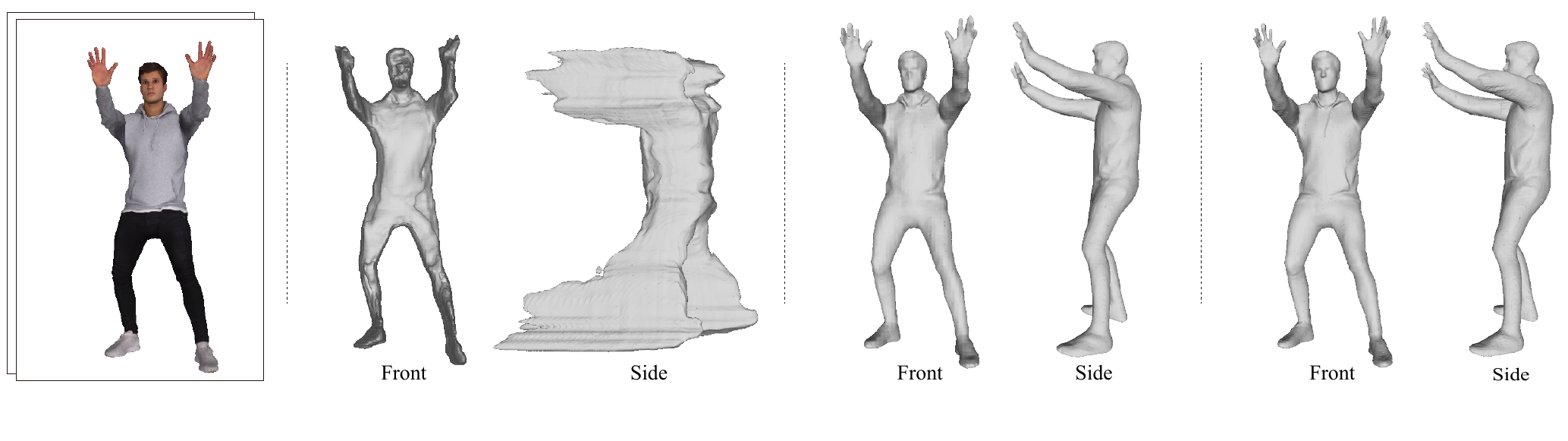}
        \put(8, 1){\footnotesize Input}
        \put(25, 1){\footnotesize PF using 2-view images + AZ}
        \put(60, 1){\footnotesize PF + VF + AZ}
        \put(86, 1){\footnotesize PF + VF + RZ}
    \end{overpic}

    \caption{Ablation study on our StereoPIFu. `PF': Pixel-aligned Feature; `PF using 2-view images': The average of PFs from 2-view images; `VF': Voxel-aligned Features; `AZ': Absolute z-value; `RZ': Relative z-offset. `PF using 2-view images + AZ' fails to infer the plausible geometry due to the complexity and diverse spatial locations of our training data. `PF + VF + AZ' is a variant of our StereoPIFu. Compared with PIFu\cite{saito2019pifu}, only our self-designed voxel-aligned features is added as the input of the implicit function. `PF + VF + RZ' is another variant of our StereoPIFu. As relative z-offset replaces absolute z-value, finer surface details can be recovered.}
    \label{fig:ablationstudy_VF_RZ}
\end{figure*}

\subsection{Datasets}
\label{sec:datasets}
We collected $171$ and $103$ rigged human models from AXYZ~\cite{axyz} and RenderPeople~\cite{renderpeople}, respectively. They all have high-fidelity geometry and realistic texture. Then non-water-tight meshes are converted into water-tight meshes with blender~\cite{blender} for efficiently generating sampling points and their ground truth occupancy values. To construct a large-scale dataset for training, these human models are animated with Mixamo~\cite{mixamo} to generate mesh sequences with various actions and postures. Finally, we generate $68419$ meshes from $246$ human models as the training data, and $7578$ meshes from $28$ human models as the testing data.

To efficiently synthesize binocular image pairs from the generated meshes, we assume the mesh surface to be predominantly Lambertian and fix the binocular camera parameters. For a given mesh in our dataset, we randomly rotate, translate, and scale the mesh within a certain range and randomly disturb the light direction and intensity to further enhance our data. The synthetic images are rendered on-the-fly with CUDA acceleration and directly used for training. This strategy effectively improves our network's generalization ability.

\subsection{Implementation Details}
We train StereoPIFu in two steps. First we train the AANet+ for depth prediction based on Eq.~\eqref{equ:stereoloss}. Then we fix the parameters of the AANet+ and train the other parts of StereoPIFu for occupancy prediction based on Eq.~\eqref{equ:occupancyloss}.

We implemented our StereoPIFu in Pytorch~\cite{paszke2019pytorch}. Adam optimizer~\cite{kingma2014adam} ($\beta_1 = 0.9,\beta_2 = 0.999$) is used and the weight decay is set to $0.0001$. The learning rate starts at $0.0001$ and is decayed by the factor of 0.1 after every 10 epochs. We train our network on 3 NVIDIA V100 GPUs with batch size of 15. The parameter $t$ of transformation function $\psi_t(\cdot)$ is set to 50. The loss weights in Eq.~\eqref{equ:stereoloss} are set to $\lambda_1 = \lambda_2 = \lambda_3 = 1.0, \lambda_4 = \frac{2}{3},\lambda_5 = \frac{1}{3}$. The resolution of our stereo image pair is set to $576 \times 576$ and their maximum disparity range is set to $72$. Due to the downsampling operator, the sizes of confidence volume $\Psi$ and cost volume $\Phi$ are $[1, 24, 192, 192]$, $[64, 24, 96, 96]$, respectively.

\begin{figure}[htp]
    \centering
    \begin{overpic}
        [width=\linewidth]{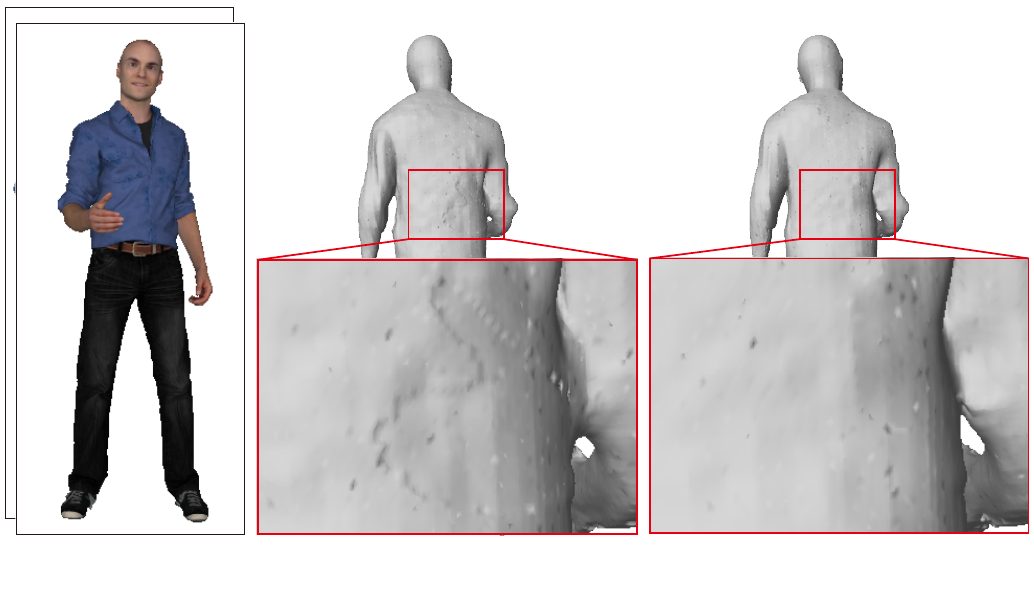}
        \put(10, 2){\footnotesize Input}
        \put(36, 2){\footnotesize wo $\psi_t(\cdot)$}
        \put(74, 2){\footnotesize with $\psi_t(\cdot)$}
    \end{overpic}
    \caption{Ablation study on transformation function $\psi_t(\cdot)$. The transformation function $\psi_t(\cdot)$ can effectively eliminate the back geometry copy artifact caused by the discontinuity of $Z_E(P)$.}
    \label{fig:ablationstudy_psi}
    \vspace{-3mm}
\end{figure}

\subsection{Evaluations}
\label{sec:evaluations}
We evaluate our StereoPIFu on three datasets, including AXYZ pose dataset~\cite{axyz}, the synthetic meshes from our testing data, and BUFF dataset~\cite{zhang2017detailed}. AXYZ pose dataset contains 100 high-quality meshes with obvious cloth wrinkles under general posture. For our testing dataset, we randomly select 100 meshes that are accompanied by some challenging postures and complex occlusions. BUFF dataset contains 100 human meshes scanned by high-end equipment. The material and lighting of these meshes look realistic. These datasets have ground-truth measurements and are not used in our training stage. The corresponding color images are obtained by using our renderer. In the following, we first show some ablation studies to analyze the role of each module in our method, and then do comparisons with other state-of-the-art methods. In the final, we show the reconstruction results on real captured data to demonstrate the generalization of our method.

\begin{table}[thp]
    \begin{center}
    \resizebox{0.70\linewidth}{!}
    {
        \begin{tabular}{|l|c|c|c|}\hline
            Method           & AXYZ & BUFF & Synthetic\\ \hline
            AANet+          & 0.7268   &	0.8969&	0.7699\\
            Retrained AANet+  & 0.1635   &	0.1785&	0.1600\\
            PF+VF+$\psi_t(\text{RZ})$  &	0.1754&	0.1907&	0.1659\\ \hline
        \end{tabular}
    }
    \end{center}
    \caption{Comparisons between AANet+ and StereoPIFu. The numerical values represent the average distance (cm) between ground truth geometry and the visible area generated by AANet+ predicting depth or our reconstructed result.}
    \label{tab:Compare_AANet_Ours}
\end{table}

\begin{table}[thp]
    \begin{center}
    \resizebox{\linewidth}{!}{
    \begin{tabular}{|l|c|c|c|c|c|c|c|c|}
    \hline
                  & \multicolumn{2}{c|}{AXYZ Pose} & \multicolumn{2}{c|}{BUFF} & \multicolumn{2}{c|}{Synthetic}\\
    \hline
    Method        & P2S & Chamfer & P2S & Chamfer & P2S & Chamfer  \\
    \hline\hline
    DeepHuman           & 2.656 &	2.670&	3.875&	3.454&	2.761&	3.502\\
    PIFu With 1-View    & 1.760 &	1.980&	2.010&	2.033&	2.729&	3.680\\
    PIFu With 2-Views   & 1.739 &	1.971&	1.975&	2.013&	2.749&	3.706\\
    PIFuHD              & 1.551 &	1.666&	1.816&	1.735&	2.544&	3.219\\
    \hline\hline
    PF+VF+AZ            & 0.612 &   0.668&  0.639&	 0.667 & 0.726 & 0.641\\
    PF+VF+RZ            & 0.556 &   0.611&	0.591&	 0.614 & 0.469 & 0.485\\
    PF+VF+$\psi_t(\text{RZ})$      
            & \textbf{0.547}&\textbf{0.603}&\textbf{0.568}&\textbf{0.612}&\textbf{0.417}&\textbf{0.436}\\
    \hline
    \end{tabular}}
    \end{center}
    \caption{Quantitative evaluation on several datasets. Point-to-Surface distance and Chamfer distance (cm) between reconstructed and ground truth meshes are computed. Please refer to Fig.~\ref{fig:ablationstudy_VF_RZ} for `PF,VF,AZ,RZ'.}
    \label{tab:Quantitativeevaluation}
    \vspace{-3mm}
\end{table}

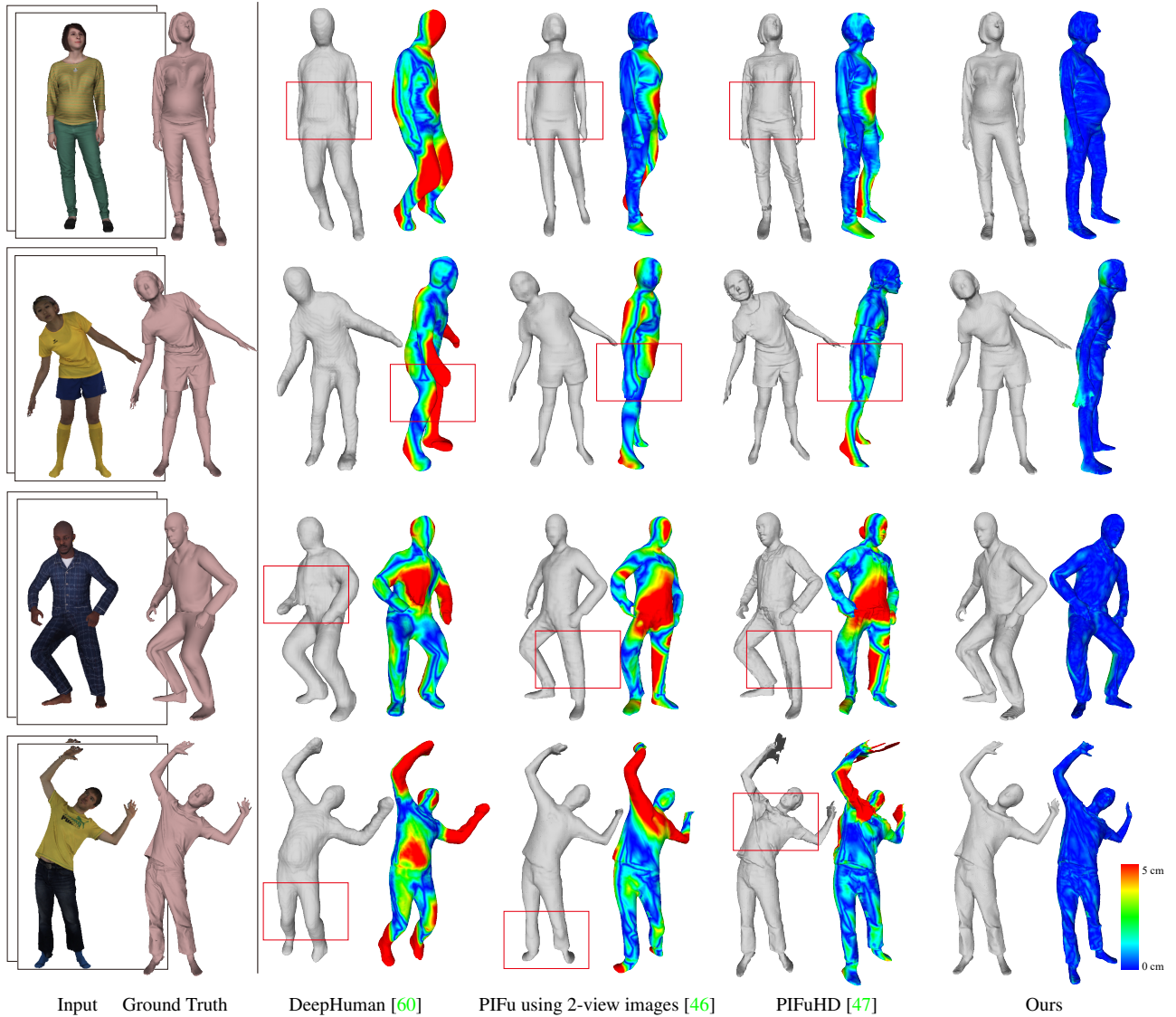
\begin{figure*}[thb]
	\centering
	\begin{overpic}
		[width=\linewidth]{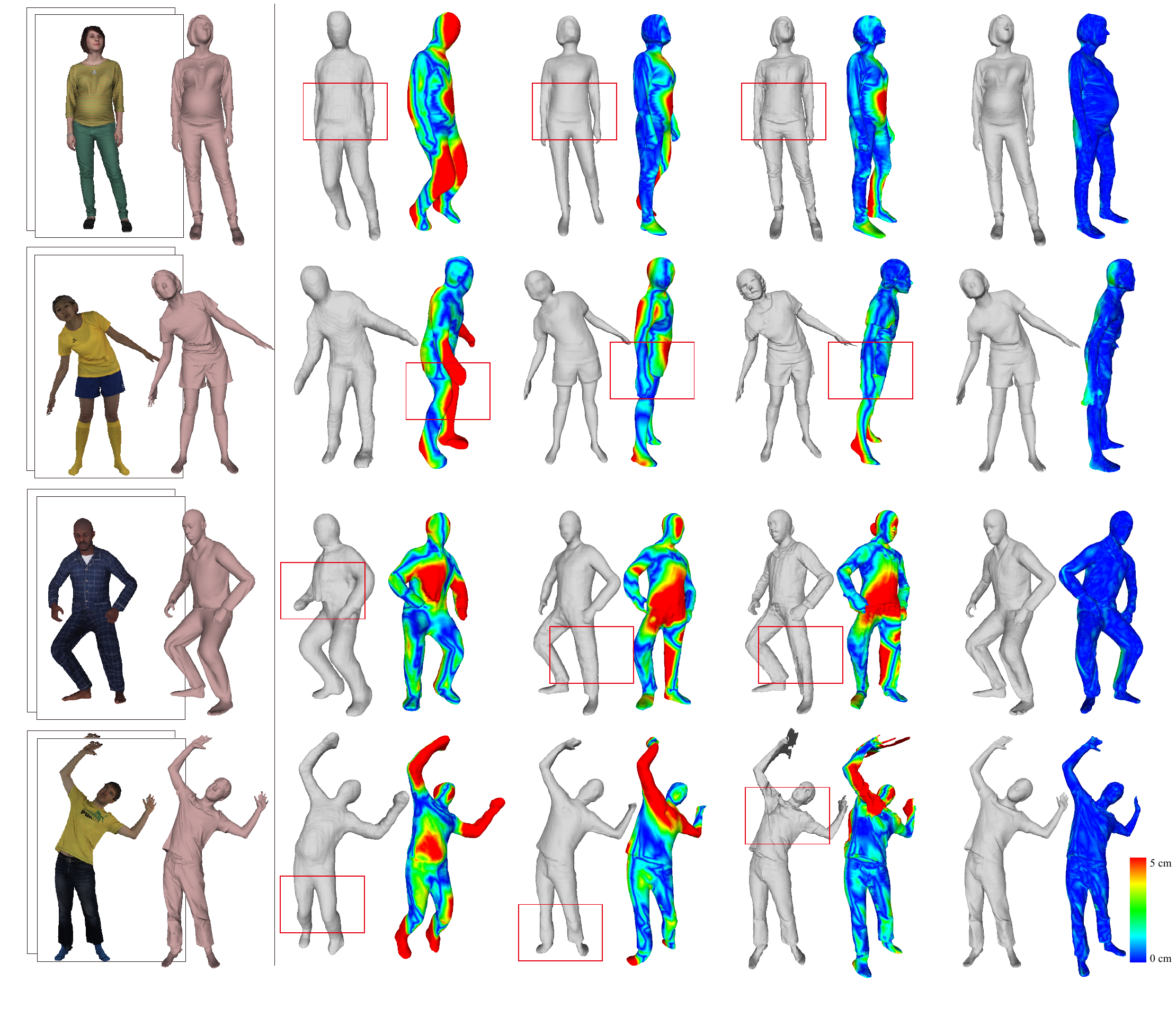}
		\put(6.5, 2){\footnotesize Input}
		\put(12, 2){\footnotesize Ground Truth}
		\put(26, 2){\footnotesize DeepHuman~\cite{zheng2019deephuman}}
		\put(42, 2){\footnotesize PIFu using 2-view images~\cite{saito2019pifu} }
		\put(67, 2){\footnotesize PIFuHD~\cite{saito2020pifuhd}}
		\put(88, 2){\footnotesize Ours}
	\end{overpic}
    \vspace*{-6mm}
	\caption{Qualitative comparison with other methods. Our StereoPIFu can depth-aware reconstruct the high-fidelity clothed human body and is more robust to extreme postures. The incorrect reconstruction parts of other methods are marked with red rectangles.}
	\label{fig:comparison}
\end{figure*}

\noindent{\bf{Ablation Study.}} We conduct ablation studies to demonstrate the importance of inputs in Eq.~\eqref{eq:functioninputs} that we design for occupancy inference and high-fidelity reconstruction. First, we retrain PIFu using 2-view images~\cite{saito2019pifu} with our constructed dataset, where PIFu's network takes the same input images as our StereoPIFu. 
As shown in Fig.~\ref{fig:ablationstudy_VF_RZ}, due to the complexity and diverse spatial locations of our training data, PIFu~\cite{saito2019pifu} using the pixel-aligned feature from 2 view images and absolute $z$-coordinate fails to reconstruct reasonable human geometry. In contrast, a variant version of our StereoPIFu successfully learns human priors from the same dataset by taking the pixel-aligned feature, voxel-aligned features, and the absolute $z$-coordinate of $P$ as input. This experiment shows that our voxel-aligned features indeed encode the query point's depth-scale information and further enhance the representation ability of previous works. 

Fig.~\ref{fig:ablationstudy_VF_RZ} also shows that geometric details can be better recovered by replacing the absolute $z$ value with the relative $z$-offset. Besides, as shown in Tab.~\ref{tab:Compare_AANet_Ours}, the errors of our StereoPIFu and retrained AANet+ are similar (slight error increase of StereoPIFu may come from the process of reconstruction from occupancy fields), which verifies that our self-designed relative $z$-offset indeed effectively utilizes the human priors from the predicted depth map to guide occupancy inference. In addition, the significant accuracy improvement of the retrained AANet+ demonstrates the effectiveness of our constructed dataset. Moreover, Fig.~\ref{fig:ablationstudy_psi} shows that our transformation function $\psi_t(\cdot)$ can help infer occluded back geometry and eliminate unreasonable artifacts.

Tab.~\ref{tab:Quantitativeevaluation} shows our quantitative evaluation using the above mentioned three types of dataset. We compute Point-to-Surface distance and Chamfer distance from the reconstructed mesh to the ground truth mesh. The results from Tab.~\ref{tab:Quantitativeevaluation} demonstrate that our novel voxel-aligned feature significantly improves the reconstruction accuracy.

\noindent{\bf{Comparisons.}} Fig.~\ref{fig:teaser}, \ref{fig:comparison} and Tab.~\ref{tab:Quantitativeevaluation} show our qualitative and quantitative comparison with state-of-the-art 3D human reconstruction methods. DeepHuman~\cite{zheng2019deephuman} first predicts a parametric human model, i.e., SMPL~\cite{SMPL:2015}, then the entire volume's occupancy value is regressed based on the predicted SMPL mesh. Its results are over-smooth due to memory limitation. Besides, incorrect body structure prediction in the first stage will result in large error in the final. PIFu's results can recover rough shape but still suffer from lacking fine-level details. PIFuHD~\cite{saito2020pifuhd} additionally uses the feature extracted from a high resolution image to guide the fine-level reconstruction. Also, it further enhances  geometric details using predicted front- and back- side normal maps. However, its results are still depth-ambiguous and cannot ensure the correctness of different parts of the human body's relative positions due to single image input. 
Tab.~\ref{tab:Quantitativeevaluation} shows the quantitative comparison between our method and these methods. Our method greatly outperforms the state-of-the-art method of PIFuHD in terms of geometric errors. On average, the Point-to-Surface distance and Chamfer distance decrease from 1.97cm to 0.51cm (74.1\% reduction) and from 2.21cm to 0.55cm (75.1\% reduction) respectively. It shows that the introduction of our well-designed voxel-aligned features can bring a substantial gain. It is worth noting that, compared with single image based methods, our method does not need to perform scaling operation before computing the reconstruction accuracy. For other methods, we need to align their results to ground-truth models based on rigid transformation with scaling.

Different from single image based methods, our method is depth-aware. As shown in the first row of Fig.~\ref{fig:comparison}, the belly of the pregnant women can be accurately recovered by our method. StereoPIFu is also aware of the bending motion and can accurately reconstruct the plausible result as shown in the second row of Fig.~\ref{fig:comparison}. Similarly, our method also successfully maintains the relative position of the right hand and the human body as shown in Fig.~\ref{fig:teaser}. As pointed out via the red rectangles, the competing methods can not handle these challenging cases. Besides, our method is more robust than other methods. For some extreme postures as shown in the bottom two rows of Fig.~\ref{fig:comparison}, our method can still obtain correct results and our reconstructions are significantly better than other methods.

\begin{figure}[thp]
    \centering
   \begin{overpic}
       [width=0.9\linewidth]{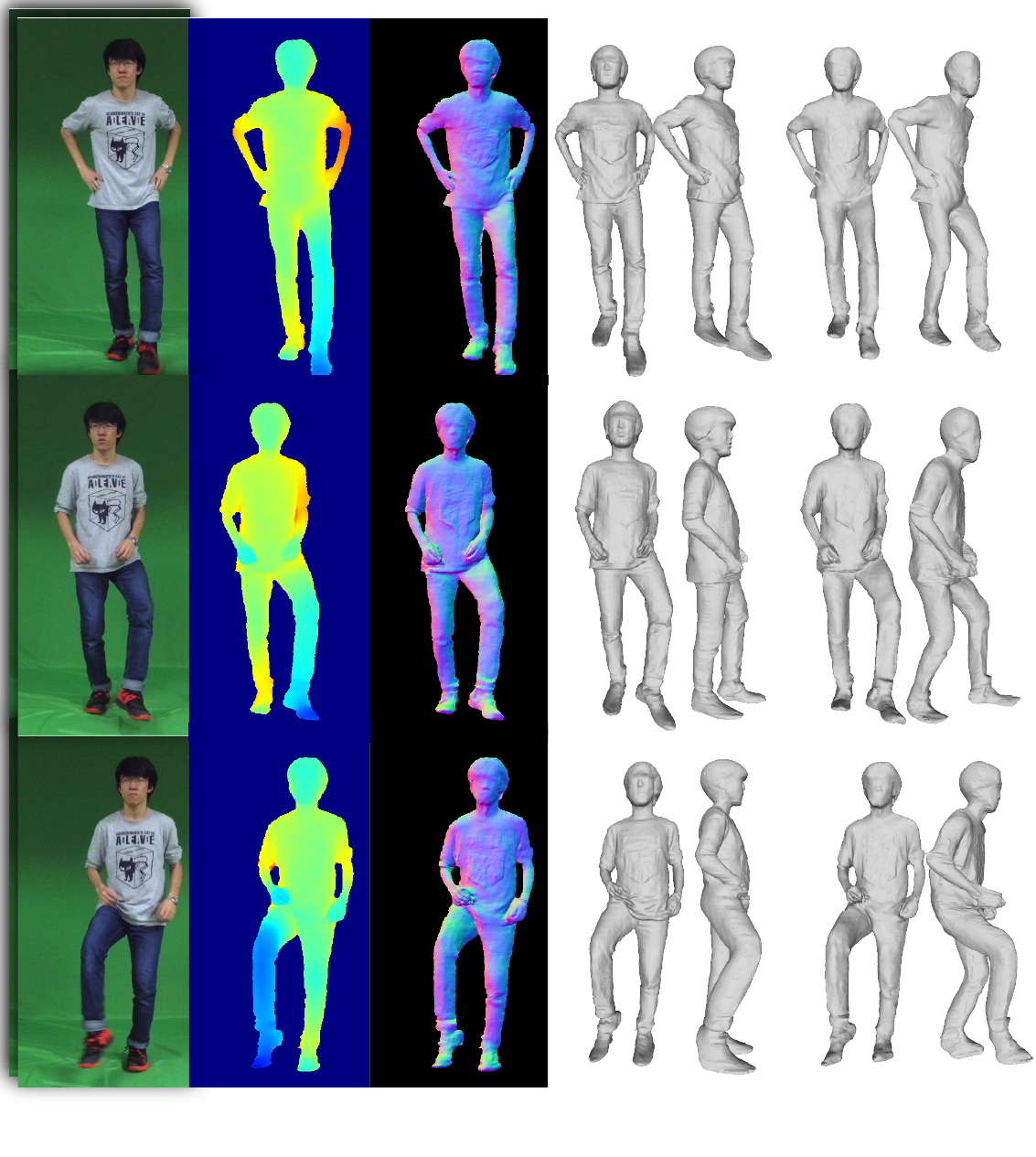}
       \put(5, 2){\footnotesize Input}
       \put(20, 2){\footnotesize Depth}
       \put(35, 2){\footnotesize Normal}
       \put(51, 2){\footnotesize PIFuHD~\cite{saito2020pifuhd}}
       \put(77, 2){\footnotesize Ours}

   \end{overpic}
    \caption{Comparison results of our method and PIFuHD~\cite{saito2020pifuhd} on real data. From left to right: input images, depth maps recovered by our network, normal maps computed from the recovered depth maps, results of PIFuHD, results of ours. Please refer to the supplementary video for sequence comparison.}
    \label{fig:realdata}
\end{figure}
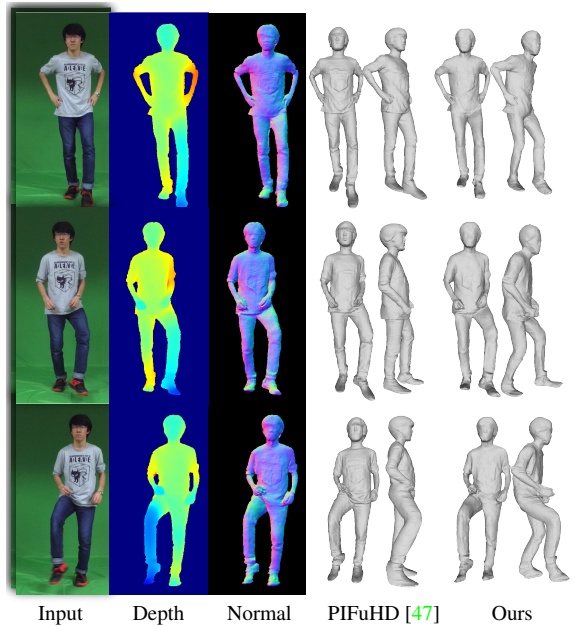

\noindent{\bf{Results on Real Data.}} We also evaluate our method on actual captured data. We use a binocular camera~\cite{ZEDCamera} to capture binocular images where people do various actions. As shown in Fig.~\ref{fig:realdata}, although the camera parameters and lighting conditions may be inconsistent with our training data, our StereoPIFu can still accurately reconstruct the geometry shape of human bodies. We also show the reconstruction results of PIFuHD~\cite{saito2020pifuhd} using a single view image. Although their results look quite good from the input view, inaccurate human body structure can be easily observed from another view. As shown in the 1st row of Fig.~\ref{fig:realdata}, the left leg of their result is longer than the right one. Besides, we can find that PIFuHD generates incorrect shape especially for the leg as shown in the last two rows, a significantly enlarged right leg can be observed from the side view of the third example. In contrast, with the help of binocular images and the well-designed neural network, our StereoPIFu can accurately recover the human body's geometry shape and relative positions of different body parts. Therefore, our method can be directly extended to capture human performance, and a comparison video is supplied as supplementary material. As the video shows, our results are more stable and robust than others.

%% file: Conclusion.tex
Our work still has several limitations. First, the careful calibration of the binocular camera is essential for our method. When the camera parameters are not accurate, our method may be affected or even fail to reconstruct body shapes. Second, for the invisible area, our StereoPIFu can only predict a plausible result while can not guarantee its accuracy. In the future, we plan to utilize several MVS systems~\cite{yao2018mvsnet,yao2019recurrent, murez2020atlas} to alleviate the problem.

In this paper, we proposed StereoPIFu, a novel clothed human digitization method that integrates stereo vision to implicit function representation. First, we introduced the novel voxel-aligned features, which enables our StereoPIFu to depth-aware clothed human body reconstruction. Second, the transformed relative z-offset of the query point is used to recover the geometric details and eliminate the back region's geometry-copy artifacts, and it further improves the reconstruction accuracy of our method. Extensive experiments demonstrate that the proposed method outperforms existing state-of-the-art methods and achieves more robust and accurate 3D human digitization.

\noindent{\bf{Acknowledgements}.} This work was supported by the Youth Innovation Promotion Association CAS (No. 2018495) and the Fundamental Research Funds for the Central Universities.